# Flexible Interpretations:
# A Computational Model for Dynamic Uncertainty Assessment

S.L. Hardt

Department of Computer Science

SUNY at Buffalo, Buffalo N.Y. 14260

The investigations reported in this paper center on the process of dynamic uncertainty assessment during interpretation tasks in real domains. In particular, we are interested here in the nature of the control structure of computer programs that can support multiple interpretation and smooth transitions between them, in real time. Each step of the processing involves the interpretation of one input item and the appropriate re-establishment of the system's confidence of the correctness of its interpretation(s).

Over the last four years, my research group has been involved in the construction of knowledge intensive systems with the goal of modeling sophisticated human-like flexible interpretation, diagnostic, and problem solving skills. The inputs to these systems contain descriptions of some reality and the systems task is to classify (diagnose) the situation according to their prior knowledge about the structure and content of the task domains. The various computer programs that have been developed process input text which may contain (1) a noisy description of the current position of a ship at seas, and its sail plans for the near future (Hardt and Rosenberg 1986), (2) an ambiguous description of events in the life of a psychiatric patient (Hardt and MacFadden 1986), (3) a description of a threat letter (Hardt 1986), or (4) a counter intuitive problem in Naive Physics (Hardt 1984). Uncertainty enters these interpretation processes in various fashions. First, the input to the program may be highly unreliable either due to noise at the input channel or due to excessive irrelevant information. Second, the program may not be able to identify and disambiguate the input due to insufficient domain knowledge.

We have developed two computational schemes to deal with uncertainty during interpretation tasks. The first scheme, which is discussed at length elsewhere, (see Hardt and Rosenberg 1985,1986 and Hardt 1986), is suitable for small knowledge domains and is implemented in the computer program ERIK that interprets ill-formed ship reports in real time. (ERIK is installed and running since Fall 1985 at the Coast Guard base in Governers Island.) In this scheme, the program pursues a single hypothesis at a time but takes into account its confidence in this hypothesis. This confidence determines the willingness of the program to change its interpretation in light of new inputs.

In the second scheme, which will be discussed here, the program computes different interpretations of the situation, in parallel. Each one of the active interpretations can assess its confidence measures. The most obvious advantage of using this scheme is that it may provide a broader interpretation of the situation and helps reduce biases so that relevant new information is not missed.

This research is supported in part by the National Science Foundation under grant number MCS-8305249 and by the United State Coast Guard under grant number DTCG2683C00227.

To implement and further investigate this method of confidence assessment, we have developed the DUNE (Diagnostic Understanding of Natural Events) system architecture that organizes the knowledge around processing structures. The system was designed as a shell for expert-systems that aid diagnoses, assessment and problem solving tasks in ill-structured domains. Currently, DUNE contains sufficient knowledge to aid the psychiatric diagnosis of anxiety and affective disorders, to perform personality assessment and to solve simple problems in physics. For a more detailed discussion of the DUNE architecture see Hardt et al 1986.

The DUNE system is built out of independent processing modules called demons. A demon may perform a variety of operations on its data, and its actions are based on both its internal state and its environment. Also, it may use information about its environment, e.g. number of demons running, and their overall priorities. The various demon capabilities can be exploited in ways that allow the expert-system to operate effectively in domains that are not entirely well defined. By being able to give estimates of the likelihood of a demon triggering the system can give the user an updated assessment of the situation based on inputs given so far. In addition, by calculating the most important condition yet to be resolved for the most confident demon in the system, the system questions the user for further information needed to resolve the present situation. The following diagram shows the architecture of the system.

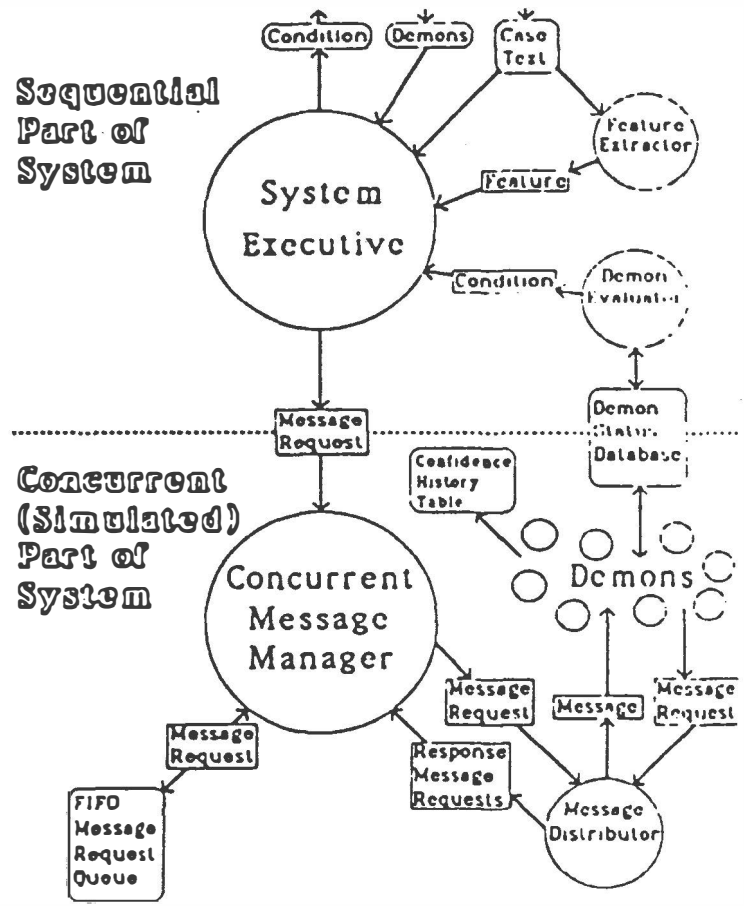

The demon's definition characteristics are as follows:
**criteria** -- A logical expression of features that determines the demons confidence.
**death_thresh** -- A confidence value threshold that is used to remove the demon from processing.
**reject_thresh** -- A confidence value threshold that is used to cause the demon to assert that it is not characterized by the features.
**accept_thresh** -- A confidence value threshold that is used to cause the demon to assert that it is characterized by the features.
**code** -- The control structure of the demon's processing (defaults to standard data demon control).

The demon's state characteristics are as follows:
**confidence** -- The numerical value which indicates the degree that the demon determines that it is characteristic of the given features.
**recvd_features** -- A list of features the demon has encountered so far.
**prev_or_bonus** -- A value used to help implementation of incremental feature processing.

Two sample runs of DUNE in the domain of psychiatric diagnosis are presented below. In these examples DUNE contains six demons processing the input concurrently. Each of the demons represents a unique psychiatric disorder. The first example which is presented in great details, shows how the system gradually centers its attention on a single diagnosis. At each step of the computation we display a table which contains among other things, the various demons (interpretations) and their confidences. In this example, the table contains the following columns (all quantities are expressed in %): the demon name, present demon states, present demon confidence, previous demon confidence, demon threshold, demon 'satisfied' threshold, demon 'not-satisfied' threshold, demon last message received numbers, demon 'raw' (uninhibited) reaction to the feature. For the sake of clarity, many details were omitted from this presentation. A fuller explanation of these concepts can be found in Hardt et al 1986.

```
input1: fatigue

DEMON                STATE   CONF  OLD  DEATH  ACCPT  REJCT  FNUM  REACT  OR-BNS
bipolar_mixed_ep     ALIVE   2     0    0      90     0      1     2      0
manic_ep             ALIVE   0     0    0      90     0      1     0      0
cyclothymic_hyp_ep   ALIVE   0     0    0      90     0      1     0      0
cyclothymic_dep_ep   ALIVE   2     0    0      90     0      1     2      0
dysthymic_ep         ALIVE   4     0    0      90     0      1     4      0
depressive_ep        ALIVE   3     0    0      90     0      1     3      0

input2: talkative

bipolar_mixed_ep     ALIVE   4     2    0      90     0      2     2      0
manic_ep             ALIVE   5     0    0      90     0      2     5      0
cyclothymic_hyp_ep   ALIVE   2     0    0      90     0      2     2      0
cyclothymic_dep_ep   ALIVE   2     0    0      90     0      2     0      0
dysthymic_ep         ALIVE   4     0    0      90     0      2     0      0
depressive_ep        ALIVE   3     0    0      90     0      2     0      0

input3: prom_dysphoric_mood

bipolar_mixed_ep     ALIVE   29    4    0      90     0      3     5      20
manic_ep             ALIVE   5     0    0      90     0      3     0      0
cyclothymic_hyp_ep   ALIVE   2     0    0      90     0      3     0      0
cyclothymic_dep_ep   ALIVE   2     0    0      90     0      3     0      0
dysthymic_ep         ALIVE   44    4    0      90     0      3     5      35
depressive_ep        ALIVE   53    3    0      90     0      3     5      45

input4: pessimistic
```

111

```
            bipolar_mixed_ep      ALIVE    31   29    0   90   0   4   2   0
            manic_ep              ALIVE     5    0    0   90   0   4   0   0
            cyclothymic_hyp_ep    ALIVE     2    0    0   90   0   4   0   0
            cyclothymic_dep_ep    ALIVE     4    2    0   90   0   4   2   0
            dysthymic_ep          ALIVE    48   44    0   90   0   4   4   0
            depressive_ep         ALIVE    56   53    0   90   0   4   3   0

            input5: distractive

            bipolar_mixed_ep      ALIVE    33   31    0   90   0   5   2   0
            manic_ep              ALIVE    10    5    0   90   0   5   5   0
            cyclothymic_hyp_ep    ALIVE     2    0    0   90   0   5   0   0
            cyclothymic_dep_ep    ALIVE     4    2    0   90   0   5   0   0
            dysthymic_ep          ALIVE    48   44    0   90   0   5   0   0
            depressive_ep         ALIVE    56   53    0   90   0   5   0   0

            input6: restless

            bipolar_mixed_ep      ALIVE    54   33    0   90   0   6   2  19
            manic_ep              ALIVE    50   10    0   90   0   6   5  35
            cyclothymic_hyp_ep    ALIVE     4    2    0   90   0   6   2   0
            cyclothymic_dep_ep    ALIVE     4    2    0   90   0   6   0   0
            dysthymic_ep          ALIVE    80   48    0   90   0   6   4  28
            depressive_ep         ALIVE    56   53    0   90   0   6   0   0

            input7: lethargic

            bipolar_mixed_ep      ALIVE    54   33    0   90   0   7   0   0
            manic_ep              ALIVE    50   10    0   90   0   7   0   0
            cyclothymic_hyp_ep    ALIVE     4    2    0   90   0   7   0   0
            cyclothymic_dep_ep    ALIVE    25    4    0   90   0   7   2  19
            dysthymic_ep          ALIVE    80   48    0   90   0   7   0   0
            depressive_ep         ALIVE    56   53    0   90   0   7   0   0

            input8: weight_disorder

            bipolar_mixed_ep      ALIVE    56   54    0   90   0   8   2   0
            manic_ep              ALIVE    50   10    0   90   0   8   0   0
            cyclothymic_hyp_ep    ALIVE     4    2    0   90   0   8   0   0
            cyclothymic_dep_ep    ALIVE    25    4    0   90   0   8   0   0
            dysthymic_ep          ALIVE    80   48    0   90   0   8   0   0
            depressive_ep         ALIVE    59   56    0   90   0   8   3   0

            input9: sleep_disorder

            bipolar_mixed_ep      ALIVE    75   56    0   90   0   9   2  17
            manic_ep              ALIVE    50   10    0   90   0   9   0   0
            cyclothymic_hyp_ep    ALIVE     4    2    0   90   0   9   0   0
            cyclothymic_dep_ep    ALIVE    25    4    0   90   0   9   0   0
            dysthymic_ep          ALIVE    84   80    0   90   0   9   4   0
            depressive_ep         ALIVE   100   59    0   90   0   9   3  38

            output from demon depressive_ep: depressive_ep
```

The following table displays the incremental confidence values for each demon during the above run:

```
            bipolar_mixed_ep      2   4  28  31  33  54  54  56  75
            manic_ep              0   5   5   5  10  50  50  50  50
            cyclothymic_hyp_ep    0   2   2   2   2   4   4   4   4
            cyclothymic_dep_ep    2   2   2   4   4   4  25  25  25
            dysthymic_ep          4   4  44  48  48  80  80  80  86
            depressive_ep         3   3  53  56  56  53  56  59 100
```

In the second example, the system does not settle on a single diagnosis. In the following we show the input sequence and the resulting table of incremental confidence values.

```
1. suicidal_thoughts
2. prom_dysphoric_mood
3. alcohol_dependence
4. prom_irritable_mood
5. irritable
6. loss_interest_pleasure
7. prom_expansive_mood
8. pessimistic
9. incoherence

manic_ep           0   0   0  47  47  47  49  49  49
cyclothymic_hyp_ep 0   0   0  22  22  22  24  24  -1
cyclothymic_dep_ep 0   0   0   0   0  21  21  21  -1
dysthymic_ep       2  49  49  49  49  50  50  51  -1
depressive_ep      1  48  48  48  51  52  52  53  94
```

At any point during the diagnosis, the system does not store or manipulate an overall certainty factor. Rather, the system's overall confidence is distributed and is determined dynamically by the individual confidences of the demons.

Current research involves issues related to the organization of the demon data base, and to the various demon mutual inhibition and enhancement schemes.

**Acknowledgements:**

This research is supported in part by the National Science Foundation under grant number MCS-8305249 and by the United State Coast Guard under grant number DTCG2683C00227.

Hardt S.L. (1984). *Naive Physics and the Physics of Diffusion: or When Intuition Fails.* Research Report Number 211, Department of Computer Science, State University of New York at Buffalo.

Hardt S.L. (1986). Challenges in the Automation of Quality Reading for Clues. Submitted to: *Expert Systems in the Government.* IEEE, Washington D.C.

Hardt S.L., MacFadden D.H. (1986) Computer Assisted Psychiatric Diagnosis: Experiments in Software Design. Submitted for publication to: *Computers in Biology and Medicine.*

Hardt, S.L. MacFadden, D. Johnson, M., Thomas, T., Wroblewski, S. (1986). *The DUNE Shell Manual: Version 1.* Research Report 86-12, Department of Computer Science, State University of New York at Buffalo.

Hardt S.L. and Rosenberg J. (1985) *The ERIK Project: Final Reports and Manuals.* Research Report 85-08, Department of Computer Science, State University of New York at Buffalo.

Hardt S.L. and Rosenberg J. (1986) Developing an Expert Ship Message Interpreter: Theoretical and Practical Conclusions. *Optical Engineering, Special Issue on Artificial Intelligence,* vol. 25, number 3.

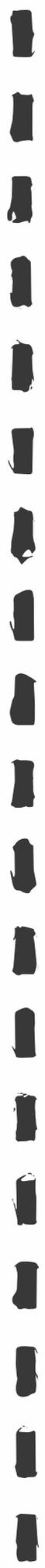